\documentclass{article}
\usepackage{arxiv}
\usepackage{multirow}
\usepackage{subcaption}
\usepackage{graphicx}
\usepackage[utf8]{inputenc} 
\usepackage[T1]{fontenc}    
\usepackage{hyperref}       
\usepackage{url}            
\usepackage{booktabs}       
\usepackage{amsfonts}       
\usepackage{nicefrac}       
\usepackage{microtype}      
\usepackage{lipsum}        
\title{PixelHop: A Successive Subspace Learning (SSL) Method 
for Object Classification}

\author{
  Yueru Chen\\
  University of Southern California\\
  Los Angeles, CA 90089 \\
  \texttt{yueruche@usc.edu} \\
   \And
  C.-C. Jay Kuo \\
  University of Southern California\\
  Los Angeles, CA 90089 \\
  \texttt{cckuo@sipi.usc.edu} \\
}

\begin{document}
\maketitle

\begin{abstract}

A new machine learning methodology, called successive subspace learning (SSL), is introduced in this work. SSL contains four key ingredients: 1) successive near-to-far neighborhood expansion; 2) unsupervised dimension reduction via subspace approximation; 3) supervised dimension reduction via label-assisted regression (LAG); and 4) feature concatenation and decision making. An image-based object classification method, called PixelHop, is proposed to illustrate the SSL design. It is shown by experimental results that the PixelHop method outperforms the classic CNN model of similar model complexity in three benchmarking datasets (MNIST, Fashion MNIST and CIFAR-10). Although SSL and deep learning (DL) have some high-level concept in common, they are fundamentally different in model formulation, the training process and training complexity. Extensive discussion on the comparison of SSL and DL is made to provide further insights into the potential of SSL.

\end{abstract}




\section{Introduction}\label{sec:intro}

Subspace methods have been widely used in signal/image processing,
pattern recognition, computer vision, etc.  \cite{turk1991eigenfaces,
bouwmans2009subspace, kriegel2009clustering, lu2011survey, gu2011joint,
wang2015joint}.  They can have different names and emphasis in various
contexts such as manifold learning \cite{wang2005adaptive,
lin2008riemannian}. Generally speaking, one uses a subspace to denote
the feature space of a certain object class, ({\em e.g.}, the subspace
of the dog object class) or the dominant feature space by dropping less
important features ({\em e.g.}, the subspace obtained via principal
component analysis or PCA).  The subspace representation offers a
powerful tool for signal analysis, modeling and processing. Subspace
learning is to find subspace models for concise data representation and
accurate decision making based on training samples. 

Most existing subspace methods are conducted in a single stage. We may
ask whether there is an advantage to perform subspace learning in
multiple stages. Research on generalizing from one-stage subspace
learning to multi-stage subspace learning is rare.  Two PCA stages are
cascaded in the PCAnet \cite{chan2015pcanet}, which provides an
empirical solution to multi-stage subspace learning. Little research on
this topic may be attributed to the fact that a straightforward cascade
of linear multi-stage subspace methods, which can be expressed as the
product of a sequence of matrices, is equivalent to a linear one-stage
subspace method. The advantage of linear multi-stage subspace methods
may not be obvious from this viewpoint. 

Yet, multi-stage subspace learning may be worthwhile under the following
two conditions.  First, the input subspace is not fixed but growing from
one stage to the other.  For example, we can take the union of a pixel
and its eight nearest neighbors to form an input space in the first
stage.  Afterward, we enlarge the neighborhood of the center pixel from
$3 \times 3$ to $5 \times 5$ in the second stage. Clearly, the first
input space is a proper subset of the second input space. By
generalizing it to multiple stages, it gives rise to a ``successive
subspace growing" process.  This process exists naturally in the
convolutional neural network (CNN) architecture, where the response in a
deeper layer has a larger receptive field. In our words, it corresponds
to an input of a larger neighborhood. Instead of analyzing these
embedded spaces independently, it is advantageous to find a
representation of a larger neighborhood using those of its constituent
neighborhoods of smaller sizes in computation and storage efficiency.
Second, special attention should be paid to the cascade interface of two
consecutive stages as elaborated below. 

When two consecutive CNN layers are in cascade, a nonlinear activation
unit is used to rectify the outputs of convolutional operations of the
first layer before they are fed to the second layer.  The importance of
nonlinear activation to the CNN performance is empirically verified, yet
little research is conducted on understanding its actual role.  Along
this line, it was pointed out in \cite{kuo2016understanding} that there
exists a sign confusion problem when two CNN layers are in cascade.  To
address this problem, Kuo {\em et al.} proposed the Saak (subspace
approximation via augmented kernels) transform \cite{kuo2018data} and
the Saab (subspace approximation via adjusted bias) transform
\cite{kuo2019interpretable} as an alternative to nonlinear activation.
Both Saak and Saab transforms are variants of PCA. They are carefully
designed to avoid sign confusion. 

One advantage of adopting Saak/Saab transforms rather than nonlinear
activation is that the CNN system is easier to explain
\cite{kuo2019interpretable}.  Specifically, Kuo {\em et al.}
\cite{kuo2019interpretable} proposed the use of multi-stage Saab
transforms to determine parameters of convolutional layers and the use
of multi-stage linear least-squares (LLS) regression to determine
parameters of fully-connected (FC) layers.  Since all parameters of CNNs
are determined in a feedforward manner without any backpropagation (BP)
in this design, it is named the ``feedforward design". Yet, the
feedforward design is drastically different from the BP-based design.
Retrospectively, the work in \cite{kuo2019interpretable} offered the
first ``successive subspace learning (SSL)" design example although the
SSL term was not explicitly introduced therein.  Although being inspired
by the deep learning (DL) framework, SSL is fundamentally different in
its model formulation, training process and training complexity. We will
conduct an in-depth comparison between DL and SSL in Sec.
\ref{sec:discussion}. 

SSL can be applied but not limited to parameters design of a CNN.  In
this work, we will examine the feedforward design as well as SSL from a
higher ground. Our current study is a sequel of cumulative research
efforts as presented in \cite{kuo2016understanding, kuo2018data,
kuo2019interpretable, kuo2017cnn}. Here, we introduce SSL formally and
discuss its similarities and differences with DL.  To illustrate the
flexibility and generalizability of SSL, we present an SSL-based machine
learning system for object classification. It is called the PixelHop
method. The block diagram of the PixelHop system deviates from the
standard CNN architecture completely since it is not a network any
longer.  The word ``hop'' is borrowed from graph theory. For a target
node in a graph, its immediate neighboring nodes connected by an edge
are called its one-hop neighbors.  Its neighboring nodes connected to
itself through $n$ consecutive edges via the shortest path are the
$n$-hop neighbors. The PixelHop method begins with a very localized
region; namely, a single pixel denoted by ${\bf p}$. It is called the
$0$-hop input.  We concatenate the attributes of a pixel, and attributes
of its one-hop neighbors to form a one-hop neighborhood denoted by
$\mathfrak{N}_1 ({\bf p})$.  We can keep enlarging the input by
including larger neighborhood regions.  This idea applies to structured
data ({\em e.g.,} images) as well as unstructured data ({\em e.g.}, 3D
point cloud sets).  An SSL-based 3D point cloud classification scheme,
called the PointHop method, was proposed in \cite{zhang2019pointhop}. 

If we implement the above idea in a straightforward manner, the
dimension of neighborhood $\mathfrak{N}_i ({\bf p})$, where $i=1, 2,
\cdots, I$ is the stage index, will grow very fast as $i$ becomes
larger.  To control the rapid dimension growth of $\mathfrak{N}_i ({\bf
p})$, we use the Saab transform to reduce its dimension. Since no label
is used in the Saab transform, it is an unsupervised dimension reduction
technique.  To reduce the dimension of the Saab responses at each stage
furthermore, we exploit the label of training samples to perform
supervised dimension reduction, which is implemented by a label-assisted
regression (LAG) unit. As a whole, the PixelHop method provides an
extremely rich feature set by integrating attributes from near-to-far
neighborhoods of selected spatial locations.  Finally, we adopt an
ensemble method to combine features and train a classifier, such as
the support vector machine (SVM) \cite{cortes1995support} and the random
forest (RF) \cite{breiman2001random}, to provide the ultimate
classification result.  Extensive experiments are conducted on three
datasets (namely, MNIST, Fashion MNIST and CIFAR-10 datasets) to
evaluate the performance of the PixelHop method. It is shown by
experimental results that the PixelHop method outperforms classical CNNs
of similar model complexity in classification accuracy while demanding
much lower training complexity. 

Our current work has three major contributions. First, we introduce the
SSL notion explicitly and make a thorough comparison between SSL and DL.
Second, the LAG unit using soft pseudo labels as presented in Sec.
\ref{subsec:clf} is novel.  Third, we use the PixelHop method as an
illustrative example for SSL, and conduct extensive experiments to
demonstrate its performance. 

The rest of this paper is organized as follows.  The PixelHop method is
presented in Sec. \ref{sec:pixelhop}.  Experimental results of the
PixelHop method are given in Sec.  \ref{sec:experiments}.  Comparison
between DL and SSL is discussed in Sec. \ref{sec:discussion}.  Finally,
concluding remarks are drawn and future research topics are pointed out
in Sec.  \ref{sec:conclusion}. 

\section{PixelHop Method}\label{sec:pixelhop}

We present the PixelHop method to illustrate the SSL methodology for
image-based object classification in this section.  First, we give an
overview of the whole system in Sec.  \ref{subsec:overview}. Then, we
study the properties of Saab filters that reside in each PixelHop unit
in Sec.  \ref{subsec:Saab}. Finally, we examine the label-assisted
regression (LAG) unit of the PixelHop system in Sec. \ref{subsec:clf}. 

\begin{figure*}[!htbp]
\begin{center}
\includegraphics[width=0.95\textwidth]{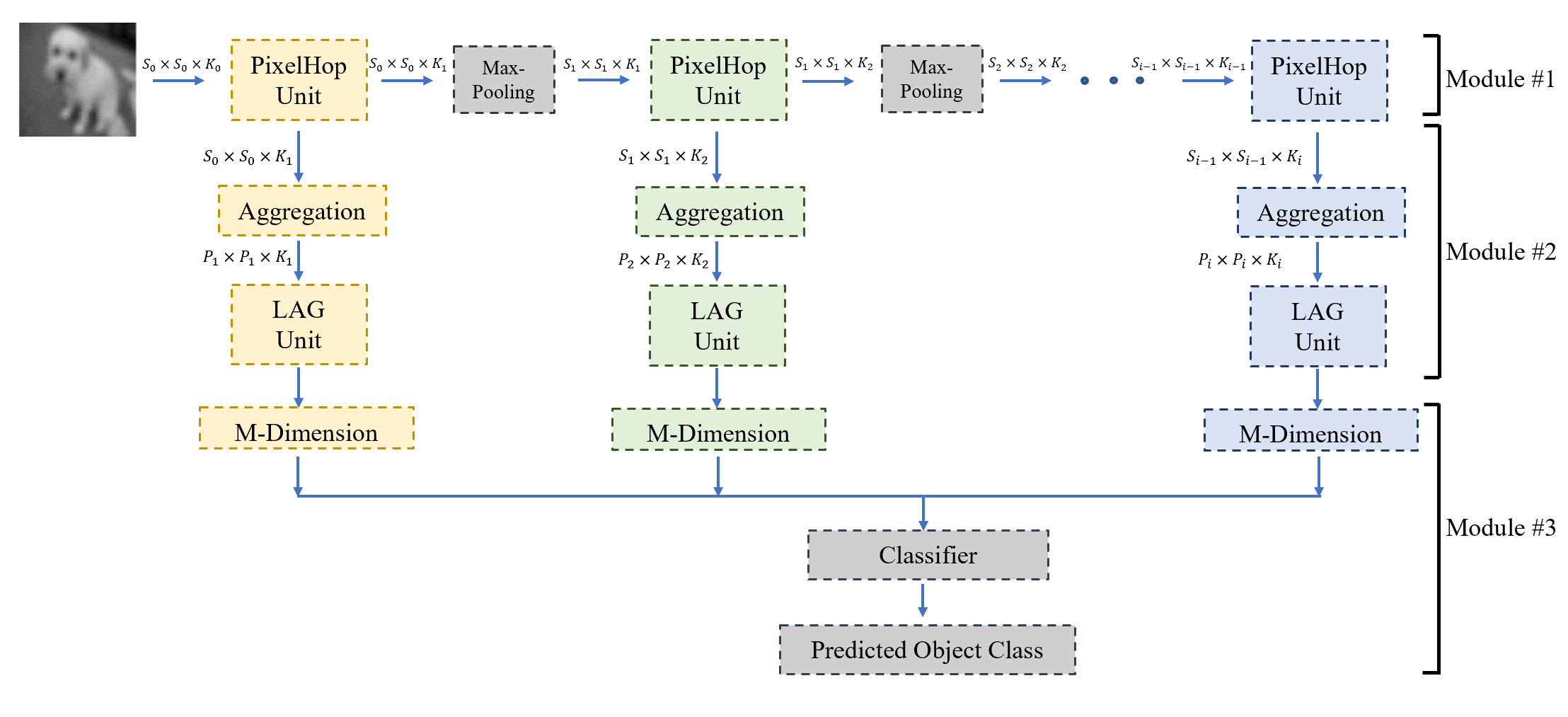}
\end{center}
\caption{The block diagram of the PixelHop method.}\label{fig:overview}
\end{figure*}

\subsection{System Overview}\label{subsec:overview}

The block diagram of the PixelHop method is given in Fig.
\ref{fig:overview}.  Its input can be graylevel or color images.  They
are fed into a sequence of $I$ PixelHop units in cascade to obtain the
attributes of the $i$th PixelHop unit, $i=1, 2, \cdots, I$, as shown in
module \#1.  The attributes in spatial locations of each PixelHop unit
are aggregated in multiple forms and, then, fed into the LAG unit for
further dimension reduction to generate $M$ attributes per unit as shown
in module \#2.  Finally, these attributes are concatenated to form the
ultimate feature vector of dimension $M \times I$ for image
classification as shown in module \#3.  The function of each module is
stated below. 

\begin{itemize}
\setlength{\itemsep}{-2pt}
\item Module \#1: A sequence of PixelHop units in cascade \\

The purpose of this module is to compute attributes of near-to-far
neighborhoods of selected pixels through $I$ PixelHop units.  The block
diagram of a PixelHop unit is shown in Fig. \ref{fig:pixelhop}.  The
$i$th PointHop unit, $i=1,\cdots, I$, concatenates attributes of the
$(i-1)$th neighborhood, whose dimension is denoted by $K_{(i-1)}$, of a
target pixel and its $N_i$ neighboring pixels to form a neighborhood
union. Through this process, the dimension of the enlarged neighborhood
is equal to $K_{(i-1)} \times (N_i+1)$.  Without loss of generality, we
set $N_i=8$ for all $i$ in our implementation. If we do not take any
further action, the attribute dimension will be $K_0 9^i$ at the $i$th
unit. It is critical to apply a dimension reduction technique so as to
control the rapidly growing dimension. This is achieved by a subspace
approximation technique; namely, the Saab transform, as illustrated in
Fig. \ref{fig:pixelhop_block}. 

Each PixelHop unit yields a neighborhood representation corresponding to
its stage index and input neighborhood size. At the $i$th PixelHop unit,
we see that the spectral dimension is reduced from $9 K_{i-1}$ to $K_i$
after the Saab transform while the spatial dimension remains the same.
Since the neighborhoods of two adjacent pixels are overlapping with each
other at each PixelHop unit, there exists spatial redundancy in the
attribute representation.  For this reason, we insert the standard
$(2\times2)$-to-$(1\times1)$ maximum pooling unit between two
consecutive PixelHop units as shown in Fig.  \ref{fig:overview}. After
the pooling, the spatial resolution is reduced from $S_{(i-1)} \times
S_{(i-1)}$ to $S_{i} \times S_{i}$. 

\begin{figure}[!htbp]
\begin{center}
\includegraphics[width=0.45\textwidth]{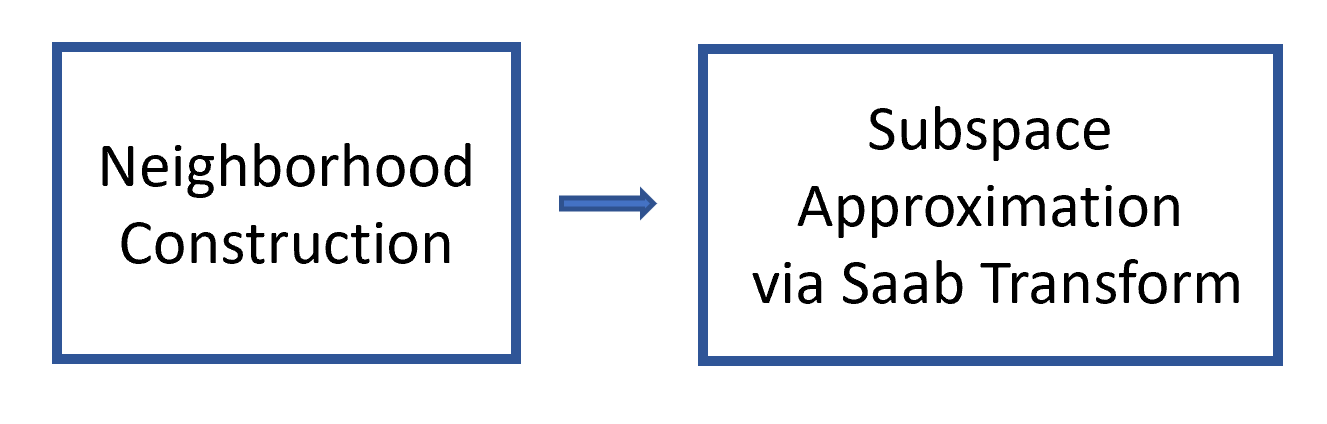}
\end{center}
\caption{The block diagram of a PixelHop unit in the PixelHop 
system.}\label{fig:pixelhop}
\end{figure}

\begin{figure*}[!htbp]
\begin{subfigure}{0.3\textwidth}
\centering
{\includegraphics[width=0.4\textwidth]{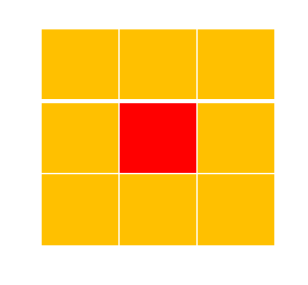}}
\caption{}
\end{subfigure}
\begin{subfigure}{0.6\textwidth}
\centering
{\includegraphics[width=1.0\textwidth]{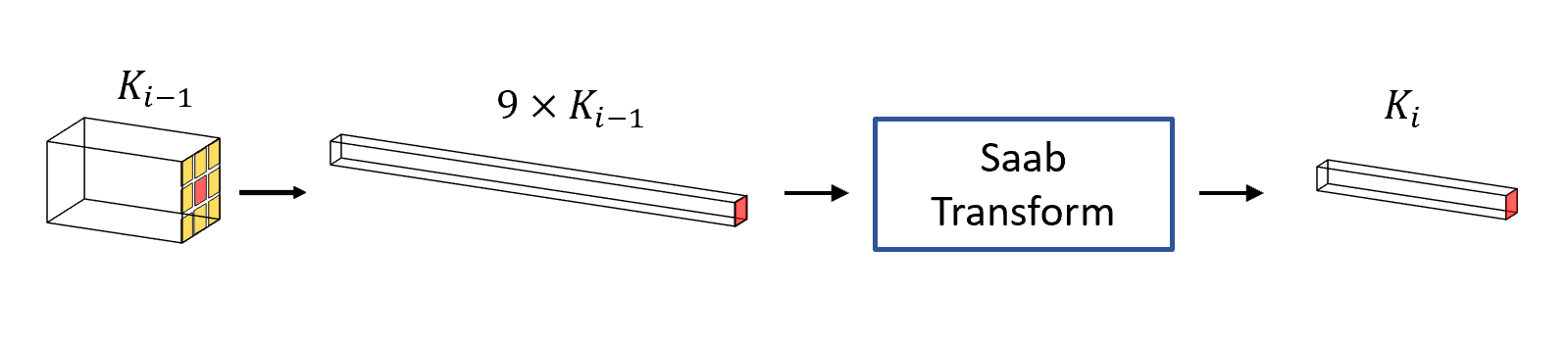}}
\caption{}
\end{subfigure}
\caption{The block diagram of a PixelHop unit: (a) neighborhood
construction by taking the union of a center pixel and its eight nearest
neighborhood pixels and (b) the use of the Saab transform to reduce the
dimension from $9 K_{(i-1)}$ to $K_i$.}\label{fig:pixelhop_block}
\end{figure*}

\item Module \#2: Aggregation and supervised dimension reduction via the
label-assisted regression (LAG) unit \\

The output from the $i$th PixelHop unit has a dimension of $S_{(i-1)}
\times S_{(i-1)} \times K_i$ as illustrated in Fig.
\ref{fig:aggregation}. The maximum pooling scheme is used to reduce the
spatial dimension before the output is fed into the next PixelHop unit
as described in module \#1.  To extract a diversified set of features at
the $i$th stage, we consider multiple aggregation schemes such as taking
the maximum, the minimum, and the mean values of responses in small
nonoverlapping regions.  The spatial size of features after aggregation
is denoted as $P_{i} \times P_{i}$, where $P_i$ is a hyper-parameter to
choose. We will explain the relationship between $P_i$ and $S_i$ in Sec.
\ref{sec:experiments}. Afterward, we reduce the feature dimension based
on supervised learning. For a given neighborhood size, we expect
attributes of different object classes to follow different distributions. 
For example, a cat image has fur texture while a car image does not. This
property can be exploited to allow us to find a more concise
representation, which will be elaborated in Sec. \ref{subsec:clf}.  In
Fig. \ref{fig:overview}, we use $M$ to denote the dimension of
the feature vector at each PixelHop unit. 

\begin{figure}[!htbp]
\begin{center}
\includegraphics[width=0.9\textwidth]{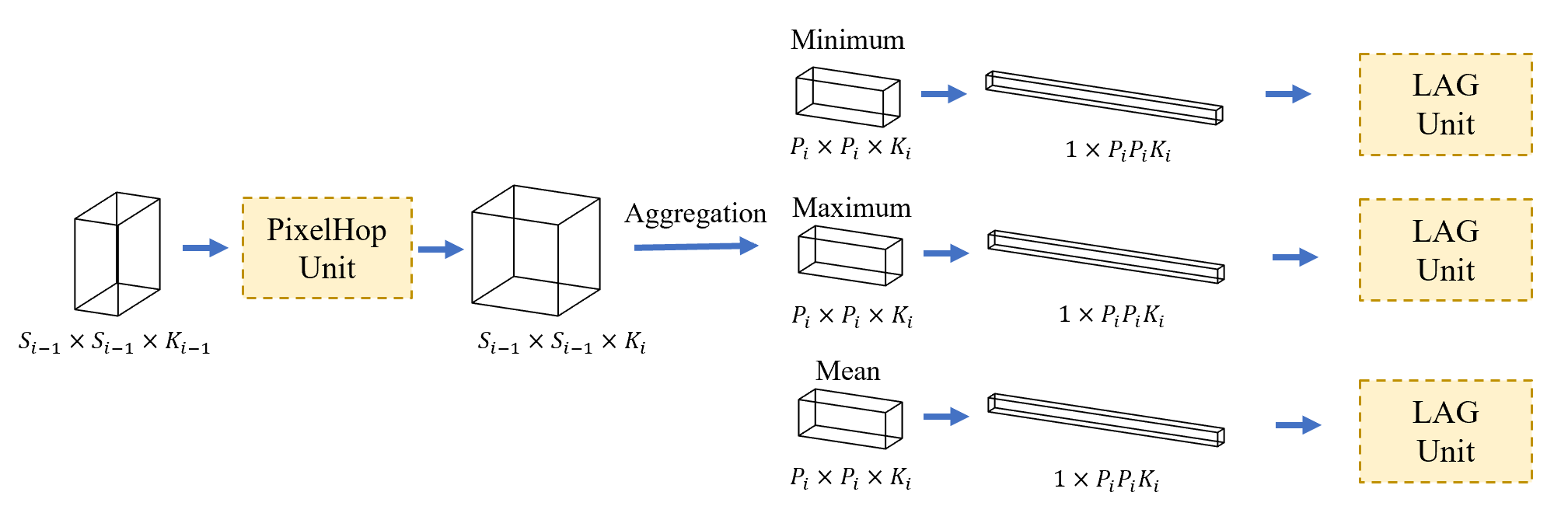}
\end{center}
\caption{Illustration of the aggregation unit in module \#2.}\label{fig:aggregation}
\end{figure}

\item Module \#3: Feature concatenation across all PixelHop units and
Classification \\ 
We concatenate $M$ features from $I$ PixelHop units to get a total of $M
\times I$ features in module \#3. Afterward, we train a multi-class
classifier using these features by following the standard pattern
recognition procedure.  In the experiment, we use the SVM classifier
with its kernel being the radial basis function (RBF). 

\end{itemize}

\subsection{Properties of Saab Filters}\label{subsec:Saab}

The Saab transform decomposes a signal space into two subspaces - the DC
(direct current) subspace and the AC (alternating current) subspace. It
uses the DC filter, which is a normalized constant-element vector, to
represent the DC subspace. It applies PCA to the AC subspace to derive
the AC filters. We will examine two issues below: 1) the relationship
between the number of AC filters and the subspace approximation
capability, and 2) the fast convergence behavior of AC filters. 

\begin{figure*}[!htbp]
\begin{center}
\begin{subfigure}[(a)]{0.35\textwidth}
\centering
{\includegraphics[width=1\textwidth,height=0.18\textheight]{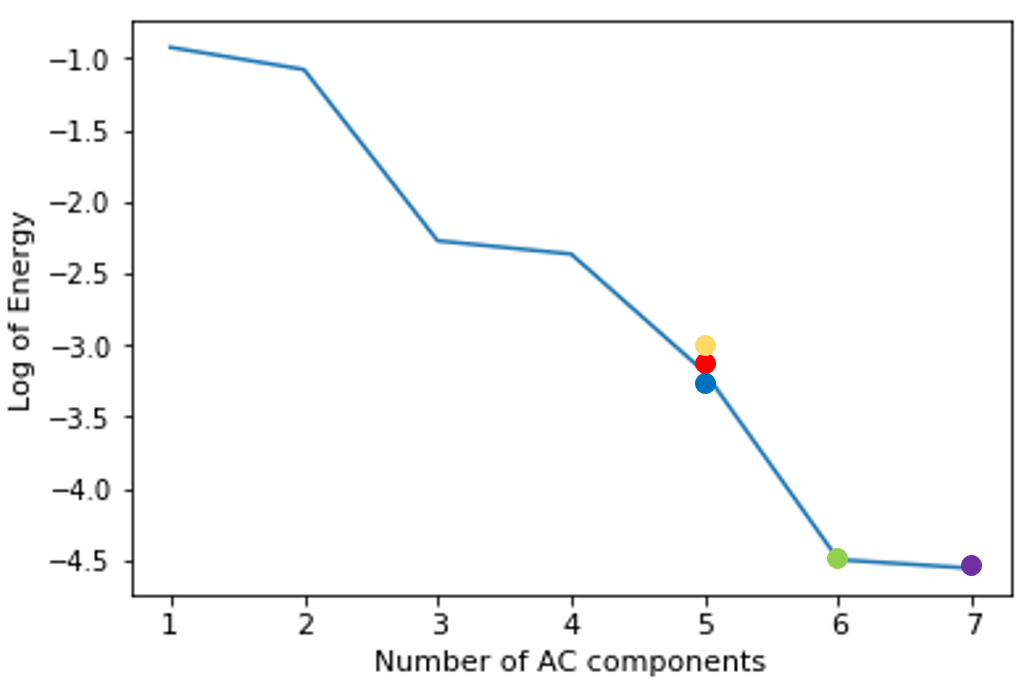}}
\caption{PixelHop Unit1}
\end{subfigure}
\begin{subfigure}[(b)]{0.35\textwidth}
\centering
{\includegraphics[width=1\textwidth,height=0.18\textheight]{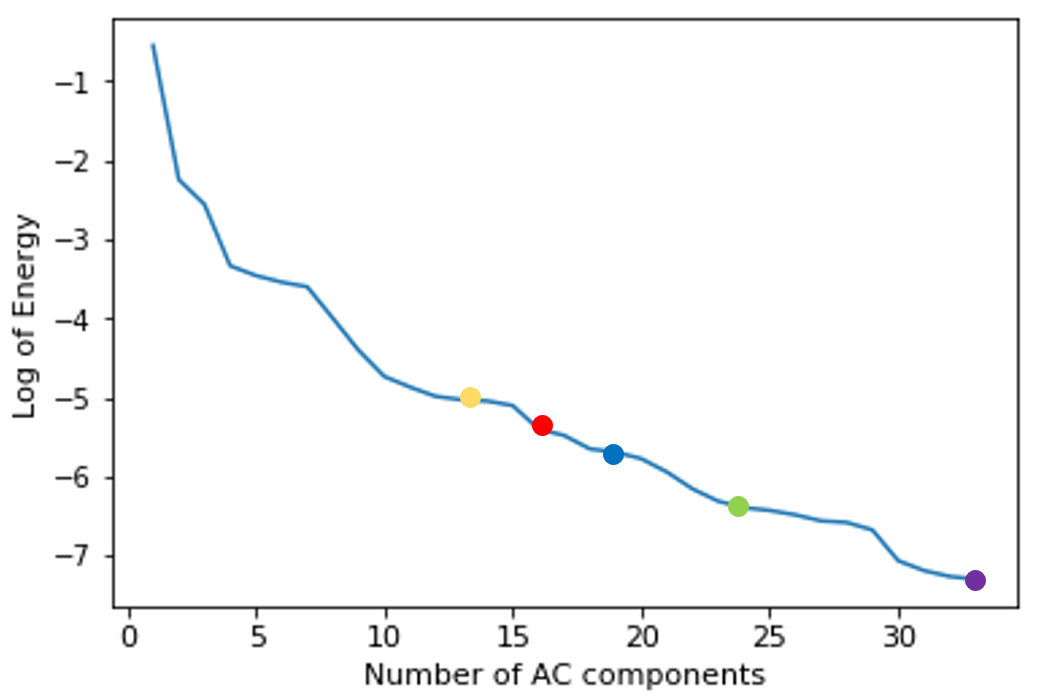}}
\caption{PixelHop Unit2}
\end{subfigure}
\begin{subfigure}[(c)]{0.35\textwidth}
\centering
{\includegraphics[width=1\textwidth,height=0.18\textheight]{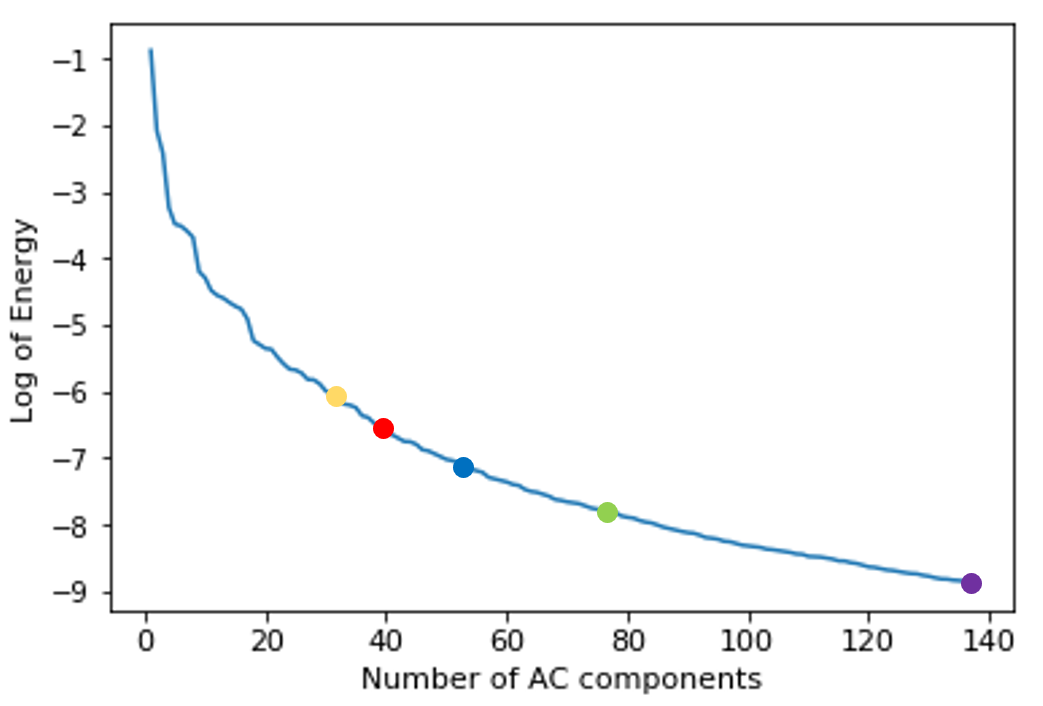}}
\caption{PixelHop Unit3}
\end{subfigure}
\begin{subfigure}[(c)]{0.35\textwidth}
\centering
{\includegraphics[width=1\textwidth,height=0.18\textheight]{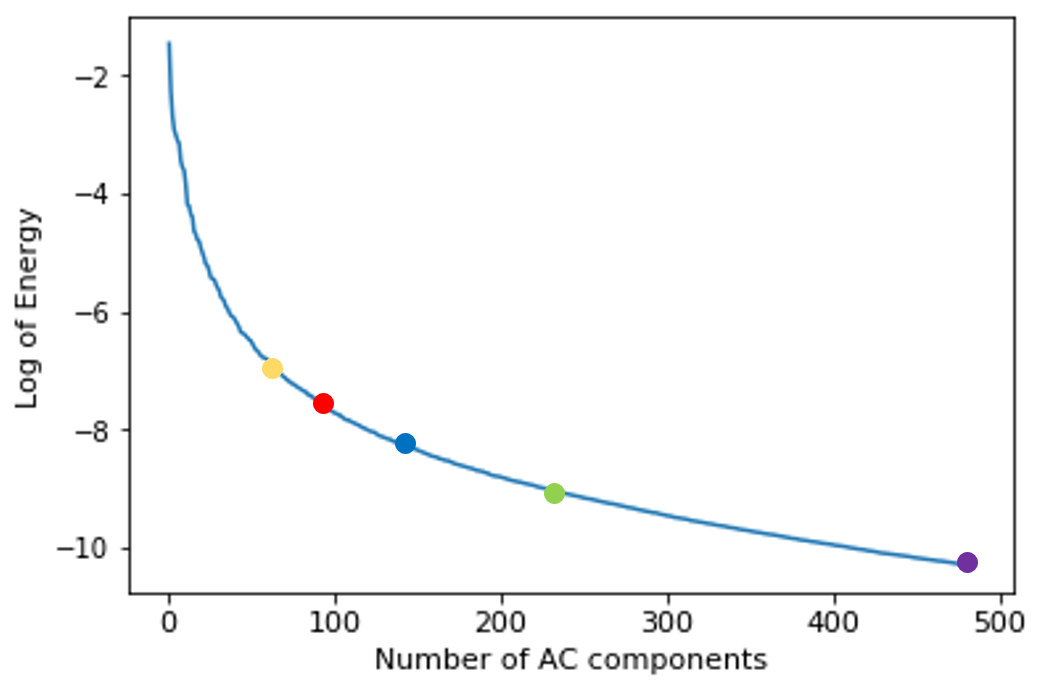}}
\caption{PixelHop Unit4}
\end{subfigure}
\end{center}
\caption{The log energy plot as a function of the number of AC filters
tested on the luminance (Y) channel of color images from CIFAR-10 dataset, where the yellow, red, blue, green
and purple dots indicate the cumulative energy ratio of 95\%, 96\%, 97\%, 98\%
and 99\%, respectively.}\label{fig:log_energy}
\end{figure*}

{\bf Number of AC filters} We show the relationship between the energy
preservation ratio and the number of Saab AC filters in Fig.
\ref{fig:log_energy}. We see that leading AC filters can capture a large
amount of energy while the capability drops rapidly as the index becomes
larger.  We plot four energy thresholds: 95\% (yellow), 96\% (red), 97\%
(blue), 98\% (green) and 99\% (purple) in Fig. \ref{fig:log_energy}.
This suggests that we may use a higher energy ratio in the beginning
PixelHop units and a lower energy ratio in the latter PixelHop units if
we would like to balance the classification performance and the
complexity. 

\begin{figure*}[!htbp]
\centering
\begin{subfigure}[(a)]{0.35\textwidth}
\centering
{\includegraphics[width=1\textwidth]{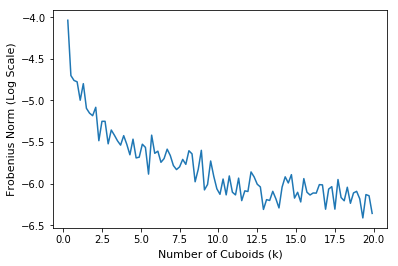}}
\caption{PixelHop Unit1}
\end{subfigure}
\begin{subfigure}[(b)]{0.35\textwidth}
\centering
{\includegraphics[width=1\textwidth]{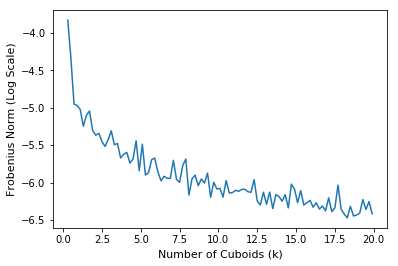}}
\caption{PixelHop Unit2}
\end{subfigure}
\begin{subfigure}[(c)]{0.35\textwidth}
\centering
{\includegraphics[width=1\textwidth]{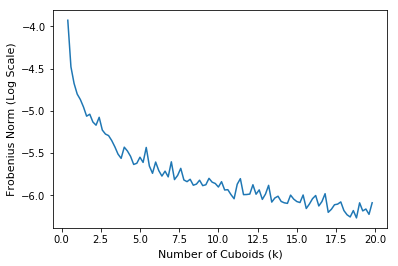}}
\caption{PixelHop Unit3}
\end{subfigure}
\begin{subfigure}[(d)]{0.35\textwidth}
\centering
{\includegraphics[width=1\textwidth]{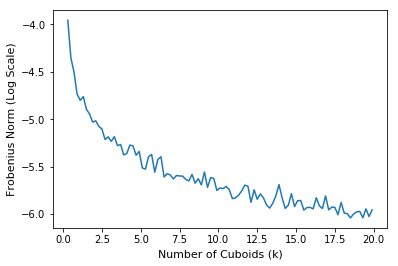}}
\caption{PixelHop Unit4}
\end{subfigure}
\caption{The Frobenius norm of the difference of two covariance 
matrices, $\Delta_{(t+1)}$, using $K_t$ and $K_{(t+1)}$ sample 
patches is plotted as a function of $K_{(t+1)}$.} \label{fig:cov_plot}
\end{figure*}

\begin{figure*}[!htbp]
\centering
\begin{subfigure}[(a)]{0.35\textwidth}
\centering
{\includegraphics[width=1\textwidth,height=0.18\textheight]{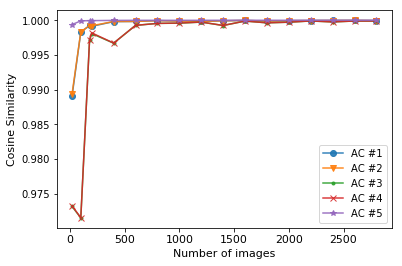}}
\caption{PixelHop Unit1}
\end{subfigure}
\begin{subfigure}[(b)]{0.35\textwidth}
\centering
{\includegraphics[width=1\textwidth,height=0.18\textheight]{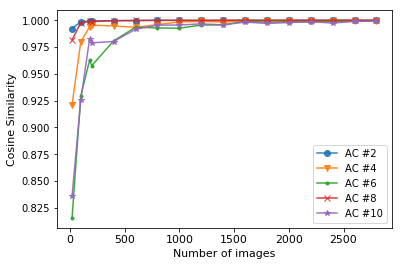}}
\caption{PixelHop Unit2}
\end{subfigure}
\begin{subfigure}[(c)]{0.35\textwidth}
\centering
{\includegraphics[width=1\textwidth,height=0.18\textheight]{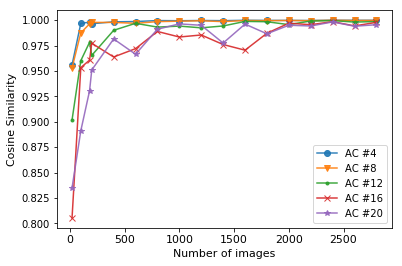}}
\caption{PixelHop Unit3}
\end{subfigure}
\begin{subfigure}[(c)]{0.35\textwidth}
\centering
{\includegraphics[width=1\textwidth,height=0.18\textheight]{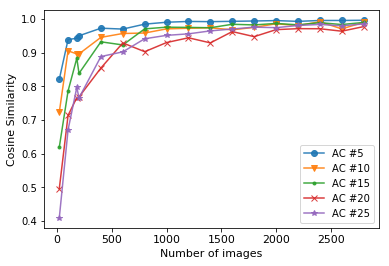}}
\caption{PixelHop Unit4}
\end{subfigure}
\caption{The cosine similarity between the AC filters obtained using all
training images and those obtained using a certain number of images is
plotted as a function of the number of images of the latter.  Five
representative AC filters at each PixelHop unit are selected for the
illustration purpose.}\label{fig:cosine_plot}
\end{figure*}

{\bf Fast Convergence.} Subspace approximation using the Saab transform
is an unsupervised learning process. The unsupervised learning pipeline
in module \# of the PixelHop system actually does not demand a large
number of data samples. We conduct experiments on the CIFAR-10 dataset
as an example to support this claim.  This system contains four PixelHop
units, where the number of AC filters is chosen by setting the energy
preservation ratio to $95\%$.  The Saab filters are derived from the
covariance matrix of DC-removed spatial-spectral cuboids. If the
covariance matrix converges quickly, the Saab filters should converge
fast as well. To check the convergence of the covariance matrix, we
compute the Frobenius norm of the difference of two covariance matrices
using $K_t$ and $K_{(t+1)}$ cuboid samples.  We plot the
dimension-normalized Frobenius norm difference, denoted by
$\Delta_{(t+1)}$, as a function of $K_{(t+1)}$ in Fig.
\ref{fig:cov_plot}. The curve is obtained as the average of five runs.
We see that the Frobenius norm difference converges to zero very
rapidly, indicating a fast-converging covariance matrix. Furthermore, we
compute the cosine similarity between the converged Saab filters (using
all 50K training images of the dataset) as well as Saab filters obtained
using a certain number of images. The results are shown in Fig.
\ref{fig:cosine_plot}. We see that AC filters converge to the final one
with about 1K training images in the first two PixelHop units and about
2.5K training images in the last two PixelHop units. 

\subsection{Label-Assisted Regression (LAG)}\label{subsec:clf}

Our design of the Label-Assisted reGression (LAG) unit is motivated by
two observations. First, each PixelHop unit offers a representation of a
neighborhood of a certain size centered at a target pixel. The size
varies from small to large ones through successive neighborhood
expansion and subspace approximation.  The representations are called
attributes.  We need to integrate the local-to-global attributes across
multiple PixelHop units at multiple selected pixels to solve the image
classification problem.  One straightforward integration is to
concatenate all attributes to form a long feature vector. Yet, the
dimension of concatenated attributes is too high to be effective. We
need another way to lower the dimension of the concatenated attributes.
Second, CNNs use data labels effectively through BP. It is desired to
find a way to use data labels in SSL. The attributes of the same object
class are expected to reside in a smaller subspace in high-dimensional
attribute space. We attempt to search for the subspace formed by samples
of the same class for dimension reduction. This procedure demands a
supervised dimension reduction technique. 

Although
being presented in different form, a similar idea was investigated in
\cite{kuo2019interpretable}. To give an interpretation to the first FC
layer of the LeNet-5, Kuo {\em et al.} \cite{kuo2019interpretable}
proposed to partition samples of each digit class into 12 clusters to
generate 12 pseudo-classes to account for intra-class variabilities.  By
mimicking the dimension of the first FC layer of the LeNet-5, we have
120 clusters ({\em i.e.}, 12 clusters per digit for 10 digits) in total.
Since each training sample belongs to one of 120 clusters, we assign it
to a one-hot vector in a space of dimension 120.  Then, we can set up a
least-squared regression (LSR) system containing 120 affine equations
that map samples in the input feature space to the output space that is
formed by 120 one-hot vectors.  The one-hot vector is used to indicate a
cluster with a hard label. In this work, we adopt soft-labeled output
vectors in setting up the LSR problem. The learning task is to use the
training samples to determine the elements in the regression matrix.
Then, we apply the learned regression matrix to testing samples for
dimension reduction. 

\begin{figure}[!htbp]
\begin{center}
\includegraphics[width=0.55\textwidth]{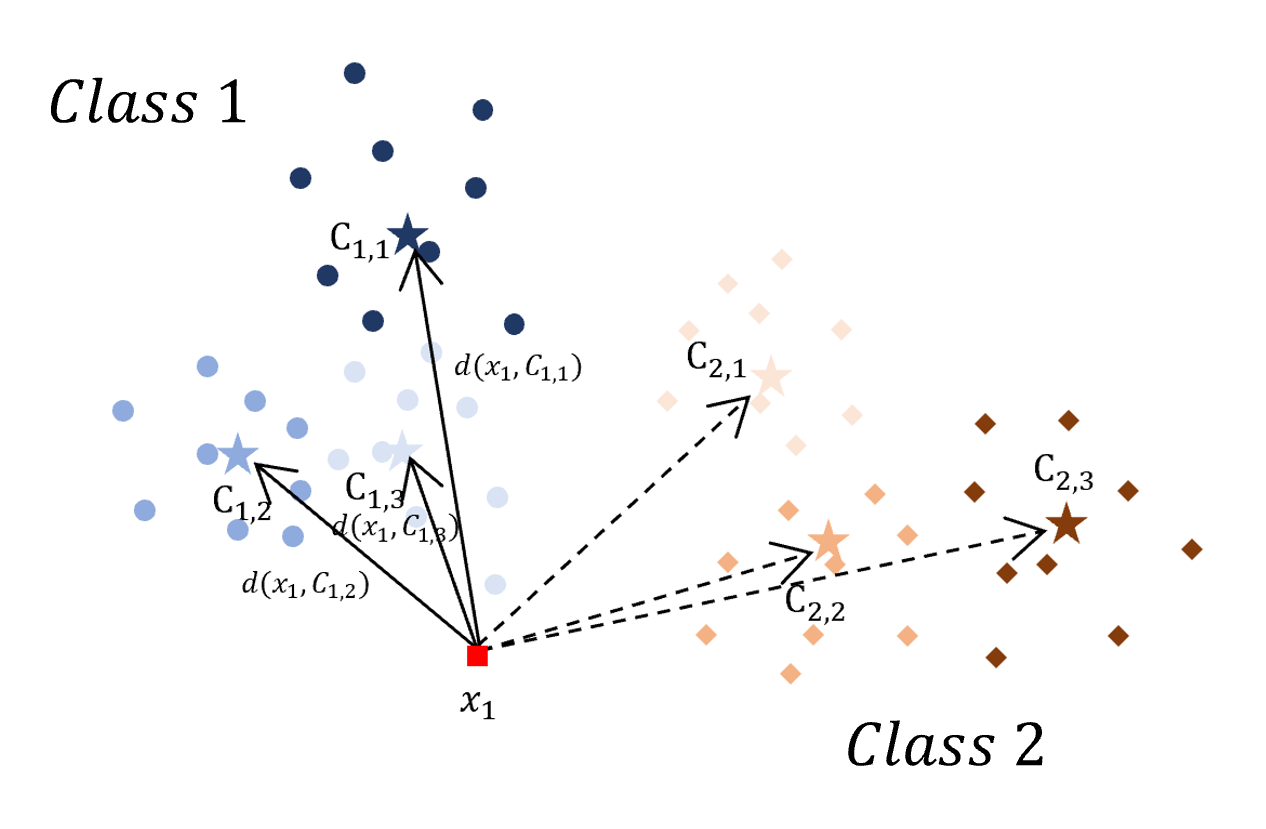}
\end{center}
\caption{Illustration of the relationship between a feature point and 
centroids of different classes in the LAG unit.}\label{fig:lag}
\end{figure}

By following the notation in Fig. \ref{fig:aggregation}, after spatial
aggregation, the $i$th PixelHop unit yields a vector of dimension $1
\times P_i P_i K_i$, where $P_i \times P_i$ denotes the number of
selected pixels and $K_i$ denotes the attribute number.  As illustrated
in Fig. \ref{fig:lag}, we study the distribution of these concatenated
attribute vectors based on their labels through the following steps:
\begin{enumerate}
\setlength{\itemsep}{-2pt}
\item We cluster samples of the same class to create object-oriented 
subspaces and find the centroid of each subspace.
\item Instead of adopting a hard association between samples and
centroids, we adopt a soft association. As a result, the target output
vector is changed from the one-hot vector to a probability vector. 
\item We set up and solve a linear LSR problem using the probability vectors.
\end{enumerate}
The regression matrix obtained in the last step is the label-assisted regressor.

For Step \#1, we adopt the k-means clustering algorithm. It applies to
samples of the same class only. We partition samples of each class into
$L$ clusters Suppose that there are $J$ object classes, denoted by
$O_j$, $j=1, \cdots, J$ and the dimension of the concatenated attribute
vectors is $n$. For Step \#2, we denote the vector of concatenated
attributes of class $O_j$ by ${\bf x}_j=(x_{j,1}, x_{j,2}, \cdots,
x_{j,n})^T \in R^n$.  Also, we denote centroids of $L$ clusters by ${\bf
c}_{j,1}$, ${\bf c}_{j,2}$, $\cdots$, ${\bf c}_{j,L}$. Then, we define
the probability vector of sample ${\bf x}_j$ belonging to centroid ${\bf
c}_{j',l}$ as
\begin{equation}\label{eq1}
\mbox{Prob}({\bf x}_j,{\bf c}_{j',l})= 0 \quad \mbox{if} j \ne j',
\end{equation}
and
\begin{equation}\label{eq2}
\mbox{Prob}({\bf x}_j,{\bf c}_{j,l})= \frac{\exp(-\alpha d({\bf x}_j,{\bf c}_{j,l}))}
{\sum_{l=1}^{L} \exp(-\alpha d( {\bf x}_j, {\bf c}_{j,l}) )},
\end{equation}
where $d({\bf x},{\bf y})$ is the Euclidean distrance between vectors
${\bf x}$ and ${\bf y}$ and $\alpha$ is a parameter to determine the
relationship between the Euclidean distance and the likelihood for a
sample belonging to a cluster. The larger $\alpha$, the probability
decays faster with the distance. The short the Euclidean distance, the
larger the likelihood. Finally, we can define the probability of sample
${\bf x}$ belonging to a subspace spanned by centroids of class $j$ as
\begin{equation}\label{eq3}
{\bf p}_{j}({\bf x}_{j'})={\bf 0}, \quad \mbox{if}  j \ne j',
\end{equation}
where ${\bf 0}$ is the zero vector of dimension $L$, and
\begin{equation}\label{eq4}
{\bf p}_{j}({\bf x}_j)=(\mbox{Prob}({\bf x}_j,{\bf c}_{j,1}), \cdots
\mbox{Prob}({\bf x}_j,{\bf c}_{j,l}), \cdots
\mbox{Prob}({\bf x}_j,{\bf c}_{j,L}))^T.
\end{equation}

Finally, we can set up a set of linear LSR equations to relate 
the input attribute vector and the output probability vector as
\begin{equation}\label{eq:l3sr}
\left[
\begin{array}{ccccc}
a_{11} & a_{12} & \cdots & a_{1n} & w_1 \\
a_{21} & a_{22} & \cdots & a_{2n} & w_2 \\
\vdots & \vdots & \ddots & \vdots & \vdots \\
a_{M1} & a_{M2} & \cdots & a_{Mn} & w_M
\end{array}
\right]
\left[\begin{array}{c}
x_{1} \\
x_{2} \\
\vdots \\
x_{n} \\
1
\end{array}
\right]
=
\left[\begin{array}{c}
{\bf p}_{1}({\bf x}) \\
\vdots \\
{\bf p}_{j}({\bf x}) \\
\vdots \\
{\bf p}_{J}({\bf x}) \\
\end{array}
\right],
\end{equation}
where $M=J \times L$ is the total number of centroids, parameters $w_1$,
$w_2$, $\cdots$, $w_M$ are the bias terms and ${\bf p}_{j}({\bf x})$ is
defined in Eq. (\ref{eq4}). It is the probability vector of dimension
$L$, which indicates the likelihood for input ${\bf x}$ to belong to the
subspace spanned by the centroids of class $j$. Since ${\bf x}$ can
belong to one class only, we have zero probability vectors with respect 
to $J-1$ classes. 

\section{Experimental Results}\label{sec:experiments}

We organize experimental results in this section as follows.  First, we
discuss the experimental setup in Sec. \ref{subsec:setup}.  Second, we
conduct the ablation study and study the effects of different parameters
on the Fashion MNIST dataset in Sec. \ref{exp:ablation}.  Third, we
perform error analysis, compare the performance of color image
classification using different color spaces and show the scalability of
the PixelHop method in Sec. \ref{subsec:error}.  Finally, we conduct
performance benchmarking between the PixelHop method and the LeNet-5
network \cite{Lecun98gradient-basedlearning}, which is a classical CNN
architecture of model complexity similar to the PixelHop method in terms
of classification accuracy and training complexity in Sec.
\ref{exp:comparison}. 

\subsection{Experiment Setup}\label{subsec:setup}

We test the classification performance of the PixelHop method on three
popular datasets: MNIST \cite{Lecun98gradient-basedlearning}, Fashion
MNIST \cite{xiao2017fashion} and CIFAR-10 \cite{krizhevsky2009learning}.
The MNIST dataset contains gray-scale images of handwritten digits (from
$0$ to $9$). It has 60,000 training images and 10,000 testing images.
The original image size is $28 \times 28$ and zero-padding is used to
enlarge the image size to $32\times32$.  The Fashion MNIST dataset
contains gray-scale fashion images. Its image size and numbers of
training and testing images are the same as those of the MNIST dataset.
The CIFAR-10 dataset has 10 object classes of color images and the image
size is $32\times32$. It has 50,000 training images and 10,000 testing
images. 

The following parameters are used in the default setting, called
PixelHop, in our experiments. 
\begin{enumerate}
\setlength{\itemsep}{-2pt}
\item Four PixelHop units are cascaded in module \#1. To decide the
number of Saab AC filters in the unsupervised dimension reduction
procedure, we set the total energy ratio preserved by AC filters to 97\%
for MNIST and Fashion MNIST and 98\% for CIFAR-10. 
\item To aggregate attributes spatially in module \#2, we average responses of
nonoverlapping patches of sizes $4 \times 4$, $4 \times 4$, $2 \times 2$
and $2 \times 2$ in the first, second, third and fourth PixelHop units,
respectively, to reduce the spatial dimension of attribute vectors.
Mathematically, we have
\begin{equation}
P_1=0.25 S_0, \quad P_2=0.25 S_1, \quad P_3=0.5 S_2, \quad P_4=0.5 S_3. 
\end{equation}
As a result, the first to the fourth PixelHop units have outputs of
dimension $8 \times 8 \times K_1$, $4 \times 4 \times K_2$, $4 \times 4
\times K_3$, and $2 \times 2 \times K_4$, respectively. Then, we feed
all of them to the supervised dimension reduction unit. 
\item We set $\alpha=10$ and the number of clusters for each object
class to $L=5$ in the LAG unit of module \#2.  Since there are $J=10$
object classes in all three datasets of concern, the dimension is
reduced to $J \times L=50$. 
\item We use the multi-class SVM classifier with the Radial Basis
Function (RBF) as the kernel in module \#3. Before training the SVM
classifier, we normalize each feature dimension to be a zero mean random
variable with the unit variance. 
\end{enumerate}
Although the hyper parameters given above are chosen empirically, the
final performance of the PixelHop system is relatively stable if their
values are in the ballpark. 

\subsection{Ablation Study}\label{exp:ablation}

\begin{table*}[h!]
\centering
\caption{Ablation study for Fashion MNIST, where the fourth and the
eighth rows are the settings adopted by PixelHop and PixelHop$^+$,
respectively.}\label{table:Ablation}
\begin{tabular}{|c|c|c|c|c|c|c|c|c|c|c|} \hline
\multicolumn{2}{|c|}{Feature Used}  & \multicolumn{2}{|c|}{DR}   
&\multicolumn{4}{|c|}{Aggregation}  
&  \multicolumn{2}{|c|}{Classifier}  
& \multirow{2}{*}{Test ACC (\%)} \\ \cline{1-10}
ALL & Last Unit & LAG &PCA&Mean&Min&Max&Skip&SVM&RF&\\ \hline
&\checkmark&\checkmark&&\checkmark&&&&\checkmark&& 89.88\\ \hline
&\checkmark&&\checkmark&\checkmark&&&&\checkmark&&89.11 \\ \hline
\checkmark&&\checkmark&&\checkmark&&&&&\checkmark&89.31 \\ \hline
\checkmark&&\checkmark&&\checkmark&&&&\checkmark&&{\bf 91.30} \\ \hline
\checkmark&&\checkmark&&&\checkmark&&&\checkmark&&91.16 \\ \hline
\checkmark&&\checkmark&&&&\checkmark&&\checkmark&&90.83 \\ \hline
\checkmark&&\checkmark&&&&&\checkmark&\checkmark&&91.14 \\ \hline
\checkmark&&\checkmark&&\checkmark&\checkmark&\checkmark&&\checkmark&&{\bf 91.68} \\ \hline
\end{tabular}
\end{table*}

We use the Fashion MNIST dataset as an example of the ablation study.
We show the test averaged classification accuracy (ACC) results for the
Fashion MNIST dataset under different settings in Table
\ref{table:Ablation}.  We can reach a classification accuracy of 91.30\%
with the default setting (see the fourth row).  It is obtained by
concatenating image representations from all four PixelHop units, using
mean-pooling to reduce the spatial dimension of attribute vectors,
label-assisted regression (LAG) and the SVM classifier.  Furthermore, we
can boost the classification accuracy by adopting three pooling schemes
({\em i.e.} max-, mean- and min-pooling) together (see the eighth row).
This is called PixelHop$^+$.

We compare the classification performance using the output of an
individual PixelHop unit, PixelHop and PixelHop$^+$ in Table
\ref{table:r_aggregation}.  We see from the table clear advantages of
aggregating features across all PixelHop units. 

\begin{table*}[h!]
\normalsize
\centering
\caption{Comparison of the classification accuracy (\%) using features 
from an individual PixelHop unit, PixelHop and PixelHop$^+$ for Fashion 
MNIST.} \label{table:r_aggregation}
\begin{tabular}{ccccccc} \hline
 Dataset& HOP-1 & HOP-2  & HOP-3 & HOP-4 & PixelHop & PixelHop$^+$ \\ \hline
 MINST &97.00 & 98.35 & 98.45 &98.71  & 98.90 &{\bf 99.09} \\ \hline
 Fashion MINST & 87.38& 89.35 & 89.96 &89.88  & 91.30 &{\bf 91.68} \\ \hline
 CIFAR-10 &52.27 & 67.86 & 69.08 &67.91  & 71.37 &{\bf 72.66} \\ \hline
\end{tabular}
\end{table*}

\begin{figure}[!htbp]
\begin{center}
\includegraphics[width=0.45\linewidth]{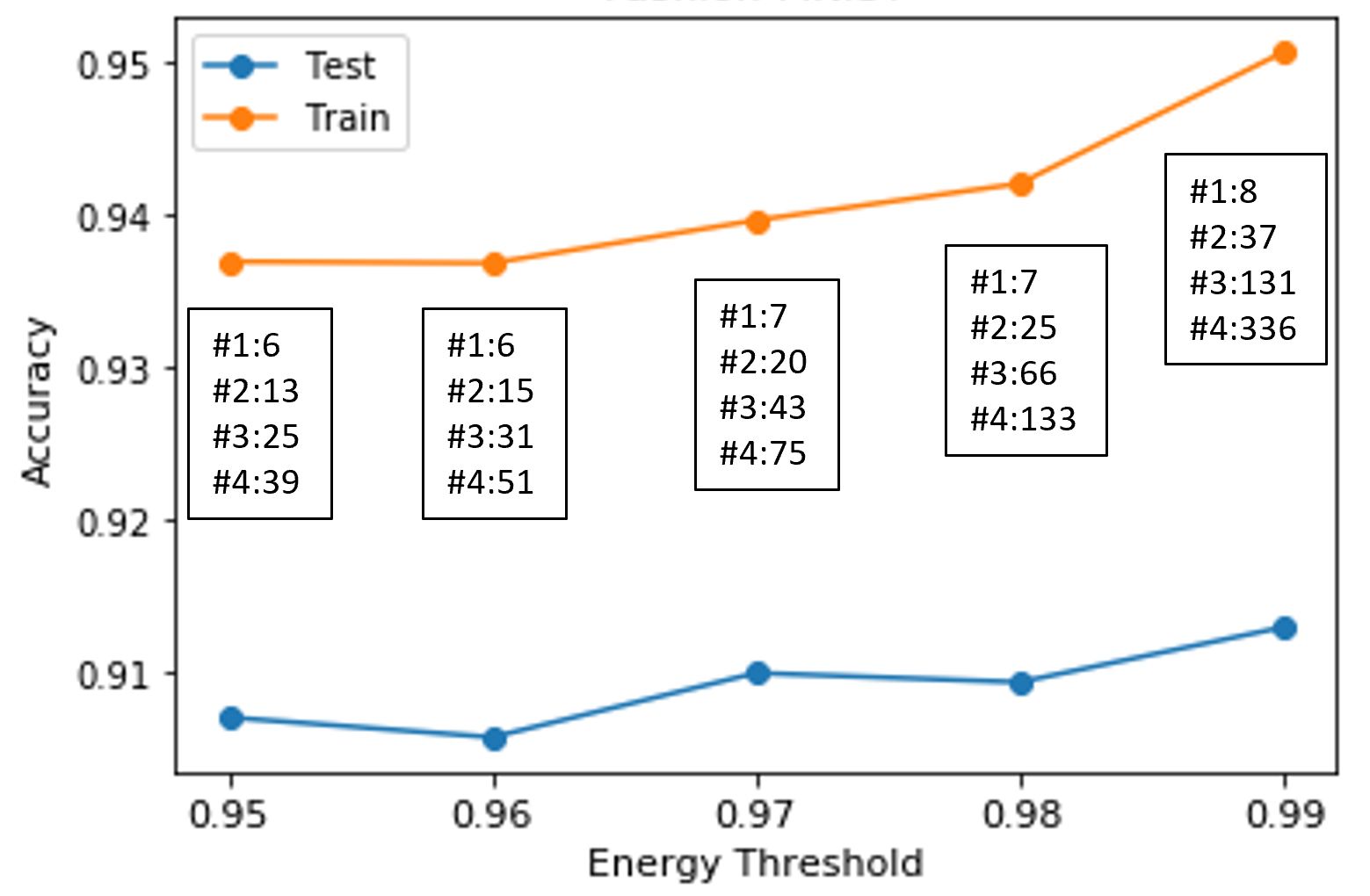}
\end{center}
\caption{The classification accuracy as a function of the total energy
preserved by AC filters tested on Fashion MNIST, where the corresponding
AC filter numbers at each PixelHop unit are also listed to illustrate the
performance-complexity tradeoff.} \label{fig:r_energy}
\end{figure}

\noindent
{\bf Number of Saab AC filters.} We study the relationship between the
classification performance and the energy preservation ratio of the Saab
AC filters in Fig.  \ref{fig:r_energy}, where the $x$-axis indicates the
cumulative energy ratio preserved by AC filters. Although preserving
more AC energy can improve the classification performance, the rate of
improvement is slow. On the other hand, we need to pay a price of adding
more AC filters. The corresponding AC filter numbers at each PixelHop
unit at each energy threshold value are listed in Fig.
\ref{fig:r_energy} to illustrate the performance-complexity tradeoff. 

\begin{table*}[h!]
\footnotesize
\centering
\caption{The confusion matrix for the MNIST dataset, where the first row shows
the predicted object labels and the first column shows the true object labels.} \label{table:cm_mnist}
\begin{tabular}{ccccccccccc} \hline
& {\bf 0} & {\bf 1} & {\bf 2} & {\bf 3} & {\bf 4} & {\bf 5} & {\bf 6} & {\bf 7} & {\bf 8} & {\bf 9} \\ \hline
{\bf 0} &0.996&    0.000&    0.000&    0.000&    0.001&    0.000&    0.001&    0.001&    0.001&    0.000 \\ \hline
{\bf 1} &0.000&    0.997&    0.001&    0.000&    0.000&    0.001&    0.001&    0.000&    0.000&    0.000 \\ \hline
{\bf 2} &0.000&    0.002&    0.992&    0.000&    0.000&    0.000&    0.000&    0.003&    0.002&    0.001 \\ \hline
{\bf 3} &0.000&    0.000&    0.002&    0.995&    0.000&    0.000&    0.000&    0.001&    0.002&    0.000 \\ \hline
{\bf 4} &0.000&    0.000&    0.003&    0.000&    0.991&    0.000&    0.001&    0.000&    0.001&    0.004 \\ \hline
{\bf 5} &0.001&    0.000&    0.000&    0.000&    0.000&    0.998&    0.001&    0.000&    0.000&    0.000 \\ \hline
{\bf 6} &0.003&    0.002&    0.000&    0.000&    0.001&    0.004&    0.987&    0.000&    0.002&    0.000 \\ \hline
{\bf 7} &0.000&    0.002&    0.008&    0.001&    0.000&    0.000&    0.000&    0.986&    0.001&    0.002 \\ \hline
{\bf 8} &0.003&    0.000&    0.004&    0.001&    0.001&    0.000&    0.000&    0.002&    0.986&    0.003 \\ \hline
{\bf 9} &0.001&    0.002&    0.003&    0.002&    0.006&    0.001&    0.000&    0.003&    0.002&    0.980 \\ \hline
\end{tabular}
\end{table*}

\begin{table*}[h!]
\scriptsize
\centering
\caption{The confusion matrix for the Fashion MNIST dataset, where the
first row shows the predicted object labels and the first column shows
the true object labels.} \label{table:cm_fmnist}
\begin{tabular}{ccccccccccc} \hline
& T-shirt/top  & Trouser& Pullover & Dress & Coat & Sandal & Shirt & Sneaker & Bag & Ankle boot \\ \hline
T-shirt/top & 0.883&    0.000&    0.015&    0.016&    0.005&    0.000&    0.072&    0.000&    0.009&    0.000 \\ \hline
Trouser & 0.001&    0.980&    0.000&    0.013&    0.002&    0.000&    0.002&    0.000&    0.002&    0.000 \\ \hline
Pullover & 0.015&    0.001&    0.877&    0.009&    0.053&    0.000&    0.044&    0.000&    0.001&    0.000 \\ \hline
Dress &0.017&    0.006&    0.010&    0.919&    0.023&    0.000&    0.022&    0.000&    0.003&    0.000 \\ \hline
Coat &0.000&    0.001&    0.056&    0.027&    0.866&    0.000&    0.050&    0.000&    0.000&    0.000 \\ \hline
Sandal &0.000&    0.000&    0.000&    0.000&    0.000&    0.979&    0.000&    0.016&    0.000&    0.005 \\ \hline
Shirt &0.110&    0.000&    0.048&    0.021&    0.072&    0.000&    0.742&    0.000&    0.007&    0.000 \\ \hline
Sneaker &0.000&    0.000&    0.000&    0.000&    0.000&    0.010&    0.000&    0.971&    0.000&    0.019 \\ \hline
Bag &0.003&    0.001&    0.003&    0.002&    0.002&    0.001&    0.001&    0.004&    0.983&    0.000 \\ \hline
Ankle boot &0.000&    0.000&    0.000&    0.000&    0.000&    0.005&    0.001&    0.026&    0.000&    0.968 \\ \hline
\end{tabular}
\end{table*}

\begin{table*}[h!]
\scriptsize
\centering
\caption{The confusion matrix for the CIFAR-10 dataset, where the first
row shows the predicted object labels and the first column shows the
true object labels.} \label{table:cm_cifar}
\begin{tabular}{ccccccccccc} \hline
& airplane & automobile& bird & cat & deer & dog & frog & horse & ship & truck \\ \hline
airplane & 0.783&    0.023&    0.034    &0.017&    0.014&    0.008&    0.012&    0.009&    0.067&    0.033 \\ \hline
automobile &0.029&    0.827&    0.010&    0.011&    0.001&    0.005&    0.007&    0.002&    0.023&    0.085 \\ \hline
bird &0.062&    0.006&    0.618&    0.064&    0.082&    0.071&    0.061&    0.016&    0.009&    0.011 \\ \hline
cat &0.023&    0.016&    0.071&    0.549&    0.061&    0.174&    0.056&    0.030&    0.008&    0.012 \\ \hline
deer &0.032&    0.003&    0.070&    0.062&    0.695&    0.031&    0.043&    0.051&    0.011&    0.002 \\ \hline
dog &0.011&    0.006&    0.059&    0.196&    0.049&    0.601&    0.026&    0.037&    0.009&    0.006 \\ \hline
frog &0.007&    0.005&    0.049&    0.059&    0.025&    0.027&    0.822&    0.002&    0.003&    0.001 \\ \hline
horse &0.023&    0.008&    0.033&    0.048&    0.052&    0.070&    0.007&    0.755&    0.001&    0.003 \\ \hline
ship &0.063&    0.042&    0.011&    0.017&    0.002&    0.006&    0.003&    0.006&    0.821&    0.029 \\ \hline
truck &0.036&    0.080&    0.010&    0.016&    0.008&    0.009&    0.005&    0.013&    0.028&    0.795 \\ \hline
\end{tabular}
\end{table*}

\subsection{Error Analysis, Color Spaces and Scalability}\label{subsec:error}

\begin{figure*}[!htbp]
\begin{center}
\includegraphics[width=0.98\textwidth]{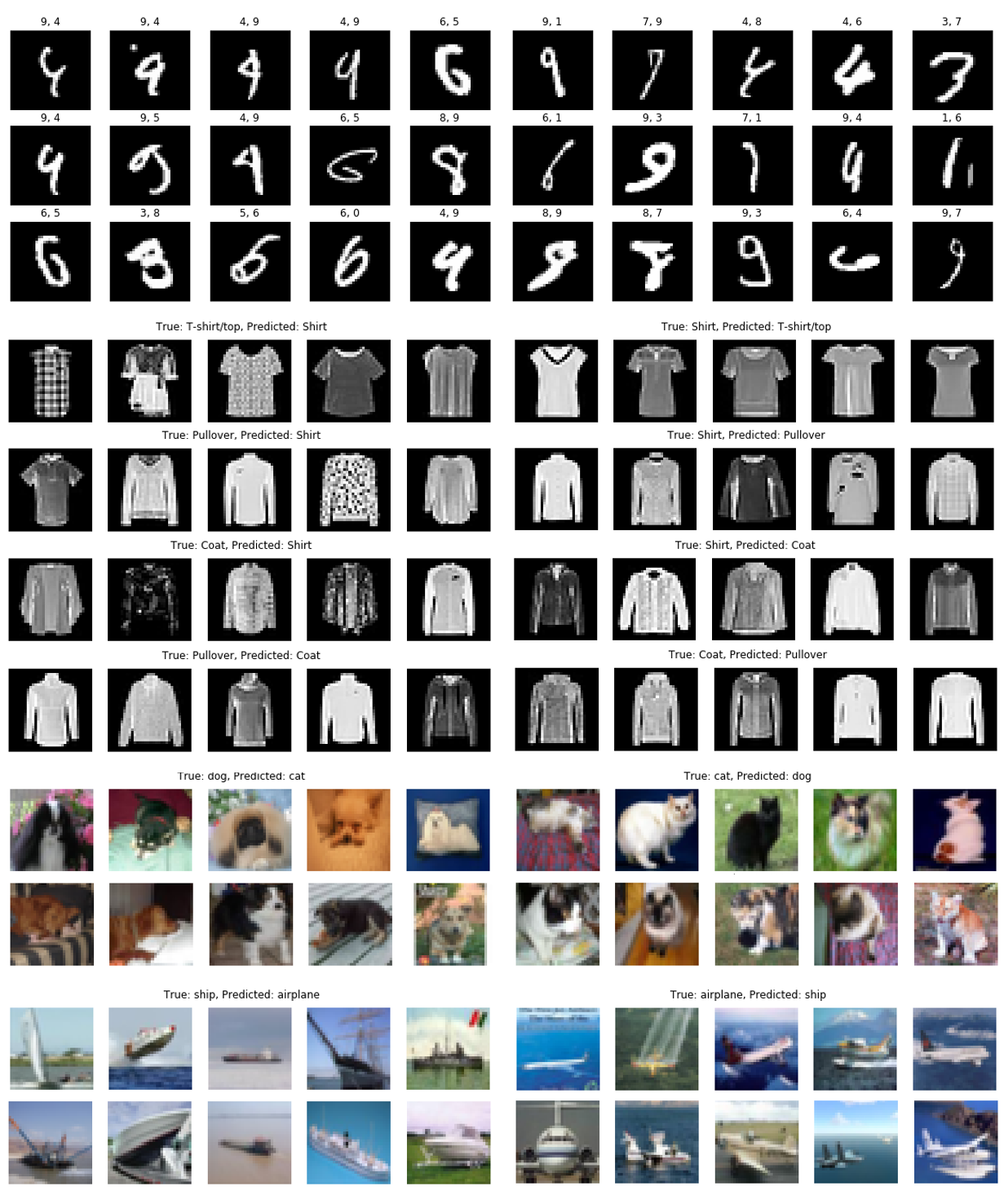}
\end{center}
\caption{Representative misclassified images in three benchmarking
datasets, where the first three rows show erroneous predictions in MNIST
and the title above each sample indicates the ground-truth and
prediction. The fourth and the seventh rows erroneous predictions of the
``shirt" class and a confusing pair ``pullover vs. coat" in Fashion
MNIST, and the last four rows show two confusing pairs, ``dog vs. cat"
and ``ship vs. airplane", in CIFAR-10.}\label{fig:error}
\end{figure*}

\noindent
{\bf Error Analysis.} We provide confusion matrices for MNIST, fashion
MNIST and CIFAR-10 in Table \ref{table:cm_mnist}, Table
\ref{table:cm_fmnist} and Table \ref{table:cm_cifar}, respectively.
Furthermore, we show some error cases in Fig. \ref{fig:error} and have
the following observations. 
\begin{itemize}
\setlength{\itemsep}{-2pt}
\item For the MNIST dataset, the misclassified samples are truly
challenging.  To handle these hard samples, we may need to turn to a
rule-based method.  For example, humans often write ``4" in two strokes
and ``9" in one stroke. If we can identify the troke number from a
static image, we can use the information to make a better prediction. 
\item For the Fashion MNIST dataset, we see that the ``shirt" class is
the most challenging one. As shown in Fig. \ref{fig:error}, the shirt
class is a general class that overlaps with the ``top", the ``pullover"
and the ``coat" classes. This is the main source of erroneous
classifications. 
\item For the CIFAR-10 dataset, the ``dog" class can be confused with
the ``cat" class.  As compared with other object classes, the ``dog" and
the ``cat" classes share more visual similarities. On one hand, it may
demand more distinctive features to differentiate them.  On the other
hand, the error is caused by the poor image resolution. The ``Ship" and
the ``airplane" classes form another confusing pair. The background is
quite similar in these two object classes, i.e. containing the blue sky
and the blue ocean. It is a challenging task to recognize small objects
in poor resolution images. 
\end{itemize}

\noindent
{\bf Different Color Spaces.} We report experimental results on CIFAR-10
with different color representations in Table \ref{table:r_color}. We
consider three color spaces - RGB, YCbCr, and Lab. The three color
channels are combined with different strategies: 1) three channels are
taken as the input jointly; 2) each channel is taken as the input
individually and all three channels are concatenated in module \#3; 3)
luminance and chrominance components are processed individually and
concatenated in module \#3. We see an advantage of processing one
luminance channel (L or Y) and two chrominance channels (CbCr or ab)
separately and then concatenate extracted features at the classification
stage. This observation is consistent with our prior experience in color
image processing. 

\begin{table}[h!]
\normalsize
\centering
\caption{Comparison of classification accuracy (\%) 
using different color representations on CIFAR-10.} 
\label{table:r_color}
\begin{tabular}{ccccccc} \hline
 & RGB &R,G,B  & YCbCr & Y,CbCr & Lab & L,ab  \\ \hline
 Testing & 68.90& 69.96 &68.74& 71.05 &67.05 & {\bf 71.37}\\ \hline
 Training &84.11 &85.06& 84.05 & 86.03 &  87.46 & {\bf87.65} \\ \hline
\end{tabular}
\end{table}

\begin{figure}[!htbp]
\centering
\begin{subfigure}[MNIST]{0.45\textwidth}
\centering
{\includegraphics[width=1\textwidth]{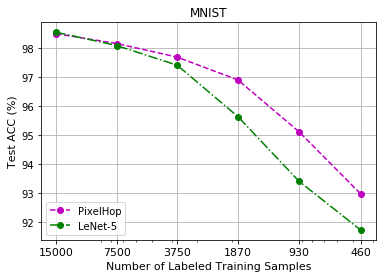}}
\end{subfigure}
\begin{subfigure}[Fashion MNIST]{0.45\textwidth}
\centering
{\includegraphics[width=1\textwidth]{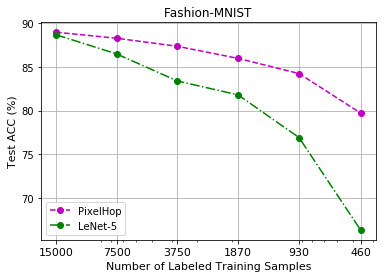}}
\end{subfigure}
\begin{subfigure}[CIFAR-10]{0.45\textwidth}
\centering
{\includegraphics[width=1\textwidth]{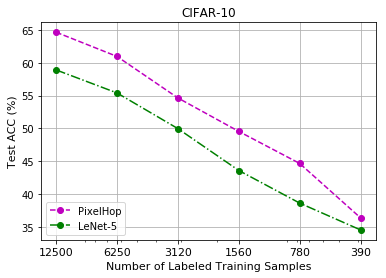}}
\end{subfigure}
\caption{Comparison of testing accuracy (\%) of LeNet-5 and
PixelHop with different training sample numbers for MNIST, 
Fashion MNIST and CIFAR-10.}\label{fig:r_Scalability}
\end{figure}

\noindent
{\bf Weak supervision.} Since PixelHop is a nonparametric learning
method, it is scalable to the number of training samples. In contrast,
LeNet-5 is a parametric learning method, and its model complexity is
fixed regardless of the training data number.  We compare the
classification accuracies of LeNet-5 and PixelHop in Fig.
\ref{fig:r_Scalability}, where only 1/4, 1/8, 1/16,1/32, 1/64, 1/128, of
the original training dataset are randomly selected as the training data
for MNIST, Fashion MNIST and CIFAR-10. After training, we apply the
learned systems to 10,000 testing data as usual.  As shown in Fig.
\ref{fig:r_Scalability}, when the number of labeled training data is
reduced, the classification performance of LeNet-5 drops faster than
PixelHop. For the extreme case with 1/128 of the original training data
size (i.e., 460 training samples), PixelHop outperforms LeNet-5 by 1.3\%
and 13.4\% in testing accuracy for MNIST and Fashion MNIST,
respectively. Clearly, PixelHop is more scalable than LeNet-5 against
the smaller training data size. This could be explained by the
fast convergence property of the Saab AC filters as discussed in
Sec. \ref{subsec:Saab}.

\subsection{Performance Benchmarking between PixelHop and LeNet-5} 
\label{exp:comparison}

We compare the training complexity and the classification performance of
the LeNet-5 and the PixelHop method in this subsection.  These two
machine learning models share similar model complexity.  We use the Lab
color space to represent images in CIFAR-10 and build two PixelHop
pipelines for the luminance and the chrominance spaces in modules \#1
and \#2 separately. Then, they are integrated in module \#3 for final
decision.  For the LeNet-5, we train all networks using TensorFlow
\cite{tensorflow2015-whitepaper} with 50 epochs and a batch size of 32.
The classic LeNet-5 architecture \cite{Lecun98gradient-basedlearning}
was designed for the MNIST dataset. We use it for the Fashion MNIST
dataset as well.  We modify the network architecture slightly to handle
color images in the CIFAR-10 dataset. The parameters of the original and
the modified LeNet-5 are shown in Table \ref{table:mLeNet-5}. The
modified LeNet-5 was originally proposed in \cite{kuo2019interpretable}. 

\begin{table}[htb]
\begin{center}
\footnotesize
\caption{Comparison of the original and the modified LeNet-5
architectures.}\label{table:mLeNet-5}
\begin{tabular}{ccc} \hline
Architecture     & Original LeNet-5  & Modified LeNet-5 \\ \hline
1st Conv Layer Kernel Size & $5 \times 5 \times 1$   &  $5 \times 5 \times 3$      \\ \hline
1st Conv Layer Kernel No.  & $6$  &  $32$     \\ \hline
2nd Conv Layer Kernel Size & $5 \times 5 \times 6$   &  $5 \times 5 \times 32$      \\ \hline
2nd Conv Layer Kernel No.  & $16$  & $64$  \\ \hline
1st FC Layer Filter No.    & $120$ & $200$ \\ \hline
2nd FC Layer Filter No.    & $84$  & $100$ \\ \hline
Output Node No.            & $10$  & $10$  \\ \hline
\end{tabular}
\end{center}
\end{table}

We compare the classification performance of the PixelHop method with
LeNet-5 and the feedforward-designed CNN (FF-CNN)
\cite{kuo2019interpretable} for all three datasets in Table
\ref{table:accuracy_1}. The FF-CNN method shares the same network
architecture with LeNet-5. Yet, it determines the model parameters in
the one-pass feedforward manner. As shown in Table
\ref{table:accuracy_1}, FF-CNN performs the worst while PixelHop$^+$
performs the best in all datasets. The latter outperforms LeNet-5 by
0.05\%, 0.6\% and 3.94\% for MNIST, Fashion MNIST and CIFAR-10,
respectively.  PixelHop also outperforms LeNet-5 for Fashion MNIST and
CIFAR-10. 

Furthermore, we compare the training time of LeNet-5 and PixelHop for
all three datasets in Table \ref{table:time}.  PixelHop takes less
training time than LeNet-5 for all three datasets in a CPU, where the
CPU is Intel(R) Xeon(R) CPU E5-2620 v3 at 2.40GHz. Although these
comparisons are still preliminary, we do see that PixelHop can be
competitive in terms of classification accuracy and training complexity. 

\begin{table}[h!]
\centering
\caption{Comparison of testing accuracy (\%) of LeNet-5,
feedforward-designed CNN (FF-CNN), PixelHop and PixelHop$^+$ for MNIST,
Fashion MNIST and CIFAR-10. } \label{table:accuracy_1}
\begin{tabular}{cccc} \hline
 Method    & MNIST  & Fashion MNIST & CIFAR-10 \\ \hline
 LeNet-5   & 99.04 & 91.08 & 68.72 \\ \hline
 FF-CNN  & 97.52 & 86.90 &  62.13 \\ \hline
 PixelHop  & 98.90 & 91.30 &  71.37 \\ \hline
 PixelHop$^+$  &{\bf 99.09} & {\bf 91.68} & {\bf 72.66} \\ \hline
\end{tabular}
\end{table}

\begin{table}[h!]
\centering
\caption{Comparison of training time of the LeNet-5 and the PixelHop method on
the MNIST, the Fashion MNIST and the CIFAR-10 datasets.} \label{table:time}
\begin{tabular}{cccc} \hline
 Method& MNIST  & Fashion MNIST & CIFAR-10 \\ \hline
 LeNet-5 & $\sim$25 min & $\sim$25 min & $\sim$45 min \\ \hline
 PixelHop &  $\sim$15 min & $\sim$15 min & $\sim$30 min \\ \hline
\end{tabular} \\
\end{table}

\section{Discussion}\label{sec:discussion}

In this section, we first summarize the key ingredients of SSL in Sec.
\ref{subsec:SSL}.  Then, extensive discussion on the comparison of SSL
and DL is made in Sec. \ref{subsec:comparison} to provide further
insights into the potential of SSL. 

\subsection{Successive Subspace Learning (SSL)}\label{subsec:SSL}

The PixelHop method presented in Sec. \ref{sec:pixelhop} offers a
concrete example of SSL.  Another design
based on the SSL principle, called the PointHop method, was proposed in
\cite{zhang2019pointhop}.  It is worthwhile to obtain a high-level
abstraction for these methods. Generally speaking, an SSL system
contains four ingredients:
\begin{enumerate}
\setlength{\itemsep}{-2pt}
\item successive near-to-far neighborhood expansion in multiple stages; 
\item unsupervised dimension reduction via subspace approximation at each stage;
\item supervised dimension reduction via label-assisted regression (LAG); and 
\item feature concatenation and decision making.
\end{enumerate}

For the first ingredient, we compute the attributes of local-to-global
neighborhoods of selected pixels in multiple stages successively. The main
advantage of this design is that the attributes of near and far
neighboring pixels can be propagated to the target pixel through local
communication. The attributes can be gray-scale pixel values (for
gray-scale images), the RGB pixel values (for color images), position
coordinates (for point cloud data sets), etc. The attributes of near
neighbors can be propagated in one hop while that of far neighbors can
be propagated through multiple hops as well as through spatial pooling.
For this reason, it is called the PixelHop (or PointHop) method.  As the
hop number becomes larger, we examine a larger neighborhood.  Yet, the
attribute dimension of a neighborhood will grow rapidly since it is
proportional to the number of pixels in the neighborhood. 

To control the speed of growing dimensions without sacrificing
representation accuracy much, we need to find an approximate subspace in
the second ingredient. Specifically, we exploit statistical correlations
between attribute vectors associated with neighborhoods. PCA is a
one-stage subspace approximation technique.  When we consider successive
subspace approximation in the SSL context, we adopt the Saab or the Saak
transform to eliminate the sign confusion problem.  This topic was
discussed in \cite{kuo2016understanding}, \cite{kuo2018data},
\cite{kuo2019interpretable}.  PCA and Saab/Saak transforms are
unsupervised dimension reduction techniques. 

In earlier (or later) stages, the neighborhood size is smaller (or
larger), the attribute vector dimension is smaller (or larger), and the
number of independent neighborhoods is larger (or smaller). All of them
can contribute to classification accuracy.  For an object class, the
attributes of its near and far neighbors follow a certain distribution.
We use the label information at all stages to achieve further dimension
reduction.  The label-assisted regression (LAG) unit was developed in
Sec. \ref{subsec:clf} for this purpose. This corresponds to the third
ingredient. 

For the last ingredient, dimension-reduced attributes from all stages
are concatenated to form the ultimate feature vector and a multi-class
classifier is trained for the classification task. 

\begin{table*}[htb]
\centering
\caption{Similarities of SSL and DL.} \label{table:similarities}
\begin{tabular}{ccc} \hline
                              & SSL                                            & DL                                    \\ \hline
Attributes collection         & Successively growing neighborhoods             & Gradually enlarged receptive fileds   \\ \hline
Attributes processing         & Trade spatial for spectral dimensions          & Trade spatial for spectral dimensions \\ \hline
Spatial dim. reduction        & Spatial pooling                                & Spatial pooling                       \\ \hline
\end{tabular}
\end{table*}

\subsection{Comparison of DL and SSL}\label{subsec:comparison} 

SSL and DL have some high-level concept in common, yet they are
fundamentally different in their models, training processes and training
complexities.  We list similarities of SSL and DL in Table
\ref{table:similarities}. 

Similarities reside in the high-level principle.  Both collect
attributes in the pixel domain by employing successively growing
neighborhoods.  Both trade spatial-domain patterns for spectral
components using convolutional filters. As the neighborhood becomes
larger, the dimension of spectral components become larger.  Due to
significant neighborhood overlapping, there exists strong redundancy
between neighborhoods of adjacent pixels.  The spatial pooling is
adopted by reducing such redundancy. 

\begin{table*}[htb]
\centering
\caption{Differences of SSL and DL.} \label{table:differences}
\begin{tabular}{ccc}\hline
                               & SSL                          & DL                \\ \hline
Model expandability            & Non-parametric model         & Parametric model  \\ \hline
Incremental learning           & Easy                         & Difficult         \\ \hline
Model architecture             & Flexible                     & Networks          \\ \hline
Model interpretability         & Easy                         & Difficult         \\ \hline
Model parameter search         & Feedforward design           & Backpropagation   \\ \hline
Training/testing complexity    & Low                          & High              \\ \hline
Spectral dim. reduction        & Subspace approximation       & Number of filters \\ \hline
Task-independent features      & Yes                          & No                \\ \hline
Multi-tasking                  & Easy                         & Difficult         \\ \hline
Incorporation of priors and constraints  & Easy               & Difficult         \\ \hline
Weak supervision               & Easy                         & Difficult         \\ \hline
Adversarial Attacks            & Difficult                    & Easy              \\ \hline
\end{tabular}
\end{table*}

Next, we show differences between SSL and DL in Table \ref{table:differences}
and elaborate them below. 
\begin{itemize}
\setlength{\itemsep}{-2pt}
\item Model expandability \\
DL is a parametric learning method.  One selects a fixed network
architecture to begin with.  The superior performance of DL is
attributed to a very large model size, where the number of model
parameters is typically larger than the number of training samples,
leading to an over-parameterized network. This could be a waste of
resource.  Traditional parametric learning methods do not have enough
model parameters to deal with datasets of a larger number of samples
with rich diversity. SSL adopts a non-parametric model.  It is flexible
to add and/or delete filters at various units depending on the size of
the input dataset. SSL can handle small and large datasets using an
expandable model. This is especially attractive for edge computing.  Its
model complexity can be adjusted flexibly based on hardware constraints
with graceful performance tradeoff. 
\item Incremental learning \\
It is challenging to adapt a trained DL model to new data classes and
samples. Since SSL employs a non-parametric model, we can check whether
existing Saab filters can express the new data well. If not, we can add
more Saab filters in the unsupervised dimension reduction part. Furthermore,
we can expand the regression matrix to accommodate new classes.
\item Model architecture \\
DL demands a network architecture that has an end node at which a cost
function has to be defined. This is essential to allow BP to train the
network parameters. In contrast, the architecture of an SSL design is
more flexible. We can extract rich features from processing units in
multiple stages. Furthermore, we can conduct ensemble learning on these
features. 
\item Model interpretability \\
The DL model is a black-box tool.  Many of its properties are not well
understood. The SSL model is a white-box, which is mathematically
transparent. 
\item Model parameter search \\
DL determines model parameters using an end-to-end optimization
approach, which is implemented by BP. SSL adopts unsupervised and
supervised dimension reduction techniques to zoom into an effective
subspace for feature extraction. The whole pipeline is conducted in a
one-pass feedforward fashion. 
\item Training and testing complexity \\
DL demands a lot of computing resources in model training due to BP. As
the number of layers goes extremely deep (say, 100 and 150 layers),
the inference can be very expensive as well. The training complexity of SSL
is significantly lower since it is a one-pass feedforward design. Its
testing complexity is determined by the stage number. If the number of
stages is small, inference can be done effectively. 
\item Spectral dimension reduction \\
Although DL and SSL both use convolutional operations, they have
different meanings. Convolutions in DL are used to transform one
representation to another aiming at end-to-end optimization of the
selected cost function for the network. Convolutions in SSL are used to
find projections onto principal components of the subspace. 
\item Task-independent features \\
DL uses both input images and output labels to determine system
parameters. The derived features are task dependent. SSL contains two
feature types: task-independent features and task-dependent features.
The features obtained by unsupervised dimension reduction are
task-independent while those obtained by supervised dimension reduction
are task-dependent. 
\item Multi-tasking \\
DL can integrate the cost functions of multiple tasks and define a new
joint cost function.  This joint cost function may not be optimal with
respect to each individual task. SSL can obtain a set of
task-independent features and feed them into different LAG units and
different classifiers to realize multi-tasking. 
\item Incorporation of priors and constraints \\
DL may add new terms to the original cost function, which corresponds to
priors and constraints. The impact of the modified cost function on the
learning system is implicit and indirect. SSL can use priors and
constraints to prune attributes of small and large neighborhoods
directly before they are fed into the classifier. 
\item Weak supervision \\
A large number of labeled data are needed to train DL models. Data
augmentation is often used to create more training samples. It was shown
in Sec. \ref{sec:experiments} that SSL outperforms DL in the weak
supervision case. This could be attributed to that the unsupervised
dimension reduction process in successive PixelHop units do not demand
labels. Labels are only needed in the LAG units and the training of a
classifier. Besides, we may adopt a smaller SSL model in the beginning.
Then, we can grow the model size by adding more confident test samples
to the training dataset.
\item Adversarial attacks \\
It is well known that one can exploit the DL network model to find a
path from the output decision space to the input data space. Then, a
decision outcome can be changed by adding small perturbations to the
input. The perturbation can be so small that humans may not be able to
see. As a result, two almost identical images will result in different
predictions. This is one major weakness of DL networks. In SSL, we
expect that weak perturbation can be easily filtered out by PCA, and it
is challenging for attackers to conduct similar attacks. 
\end{itemize}

\section{Conclusion and Future Work}\label{sec:conclusion}

A successive subspace Learning (SSL) methodology was introduced and the
PixelHop method was proposed in this work. In contrast with traditional
subspace methods, SSL examines the near- and far-neighborhoods of a set
of selected pixels. It uses the training data to learn three sets of
parameters: 1) Saab filters for unsupervised dimension reduction in the
PixelHop unit, 2) regression matrices for supervised dimension reduction
in the LAG unit, and 3) parameters required by the classifier. Extensive
experiments were conducted on MNIST, Fashion MNIST and CIFAR-10 to
demonstrate the superior performance of the PixelHop method in terms of
classification accuracy and training complexity. 

SSL is still at its infancy. There exist rich opportunities for further
development and extension. A couple of them are mentioned below. First,
generative adversarial networks (GAN) have been developed as generative
models, and they find applications in style transfer, domain adaptation,
data augmentation, etc. It seems feasible to develop an SSL-based
generative model. That is, we need to ensure that attribute vectors
associated with neighborhoods of various sizes of the source-domain and
target-domain images share the same distribution. Second, we would like
to investigate SSL-based contour/edge detection and image segmentation
techniques.  Historically, contour/edge detection and image segmentation
played an important role in low-level computer vision. Their importance
drops recently due to the flourish of DL. Yet, any feedforward computer
vision pipeline should benefit from these basic operations. Based on
this foundation, we can tackle with object detection and object
recognition problems using the SSL framework.

\section*{Acknowledgement}\label{sec:acknowledgement}

This research is supported in part by DARPA and Air Force Research
Laboratory (AFRL) under agreement number FA8750-16-2-0173 and in part by
the U.S. Army Research Laboratory's External Collaboration Initiative
(ECI) of the Director's Research Initiative (DRIA) program. The U.S.
Government is authorized to reproduce and distribute reprints for
Governmental purposes notwithstanding any copyright notation hereon.
The views and conclusions contained in this document are those of the
authors and should not be interpreted as necessarily representing the
official policies or endorsements, either expressed or implied, of
DARPA, the Air Force Research Laboratory (AFRL), the U.S.  Army Research
Laboratory (ARL) or the U.S. Government.

\bibliographystyle{unsrt}  
\bibliography{refs}

\end{document}